\documentclass[daat]{ipart}

\RequirePackage{hyperref}

\startlocaldefs
\theoremstyle{plain}

\endlocaldefs
\startlocaldefs
\theoremstyle{plain}

\endlocaldefs

\pubyear{2025}
\volume{1}
\issue{1}
\firstpage{1}
\lastpage{4}

\usepackage[utf8]{inputenc} % allow utf-8 input
\usepackage[T1]{fontenc}    % use 8-bit T1 fonts
\usepackage{hyperref}       % hyperlinks
\usepackage{url}            % simple URL typesetting
\usepackage{booktabs}       % professional-quality tables
\usepackage{amsfonts}       % blackboard math symbols
\usepackage{nicefrac}       % compact symbols for 1/2, etc.
\usepackage{microtype}      % microtypography
\usepackage{xcolor}         % colors
\usepackage{amsmath}
\usepackage{bbding}
\usepackage{array}
\usepackage{algorithm}
\usepackage{algorithmic}
\usepackage{graphicx}
\usepackage{textcomp}
\usepackage{subfigure} 
\usepackage{multirow}
\usepackage{multicol}
\usepackage{xcolor, colortbl}
\usepackage{bm}
\usepackage{bbm}
\usepackage[normalem]{ulem}
\usepackage{enumitem}
\usepackage{caption}
\usepackage{nicefrac}
\usepackage{wrapfig}

\definecolor{teagreen}{rgb}{0.82, 0.94, 0.75}
\definecolor{LightCyan}{rgb}{0.88,1,1}
\definecolor{LightGray}{gray}{0.8}
\definecolor{tearose}{rgb}{0.99, 0.87, 0.9}
\definecolor{powderblue}{rgb}{0.69, 0.88, 0.9}

\def \mL{\mathcal L}

\newcommand{\btheta}{\bm{\theta}}

\newcommand{\be}{\mathbf{e}}

\newcommand{\br}{\mathbf{r}}

\newcommand{\cT}{\mathcal{T}}
\newcommand{\cF}{\mathcal{F}}
\newcommand{\cA}{\mathcal{A}}

\newcommand{\cL}{\mathcal{L}}

\newcommand{\cS}{\mathcal{S}}
\newcommand{\cQ}{\mathcal{Q}}

\newcommand{\rnum}[1]{\uppercase\expandafter{\romannumeral#1}}

\newcommand{\ms}[2]{$#1_{(#2)}$}
\newcommand{\bms}[2]{$\mathbf{#1}_{(#2)}$}
\newcommand{\sbms}[2]{$\underline{#1}_{(#2)}$}

\newcolumntype{L}[1]{>{\raggedright\let\newline\\\arraybackslash\hspace{0pt}}m{#1}}
\newcolumntype{C}[1]{>{\centering\let\newline  \\\arraybackslash\hspace{0pt}}m{#1}}
\newcolumntype{R}[1]{>{\raggedleft\let\newline \\\arraybackslash\hspace{0pt}}m{#1}}
\newcommand{\TheName}[0]{EmerFlow}

\begin{document}

\begin{frontmatter}
	\title{LLM-Empowered Representation Learning for Emerging Item Recommendation}

\begin{aug}
	\author{\inits{Z.}\fnms{Ziying} \snm{Zhang}\ead[label=e1]{ziying-z21@mails.tsinghua.edu.cn}},
	\address{Department of Electronic Engineering, Tsinghua University\\
		China\\
		\printead{e1}}
	\author{\inits{Q.}\fnms{Quanming} \snm{Yao}\ead[label=e3]{qyaoaa@tsinghua.edu.cn}}%
	\address{Department of Electronic Engineering, Tsinghua University\\
		China\\
		\printead{e3}}
		\and
	\author{\inits{Y.}\fnms{Yaqing} \snm{Wang}\thanksref{t2}\ead[label=e2]{wangyaqing@bimsa.cn}}
	\address{Beijing Institute of Mathematical Sciences and Applications\\
	China\\
	\printead{e2}}
	\thankstext{t2}{Corresponding author.}
\end{aug}
\received{\sday{20} \smonth{10} \syear{2025}}

\begin{abstract}
In this work, we tackle the challenge of recommending emerging items, whose interactions gradually accumulate over time. Existing methods often overlook this dynamic process, typically assuming that emerging items have few or even no historical interactions. Such an assumption oversimplifies the problem, as a good model must preserve the uniqueness of emerging items while leveraging their shared patterns with established ones.
To address this challenge, we propose EmerFlow, a novel LLM-empowered representation learning framework that generates distinctive embeddings for emerging items. It first enriches the raw features of emerging items through LLM reasoning, then aligns these representations with the embedding space of the existing recommendation model. Finally, new interactions are incorporated through meta-learning to refine the embeddings. This enables EmerFlow to learn expressive embeddings for emerging items from only limited interactions.
Extensive experiments across diverse domains, including movies and pharmaceuticals, show that EmerFlow consistently outperforms existing methods.
\end{abstract}


%\tableofcontents
\end{frontmatter}

\section{Introduction}

In recommendation systems, new items continuously appear, initially without interaction records but gradually accumulating them over time. This phenomenon is commonly observed in domains such as movie recommendation~\cite{metaE,du2025perscen}, disease–gene association prediction~\cite{disease-gene}, and new drug recommendation~\cite{edge}. As these items accrue interactions, they provide richer data for improving predictive models. In this work, we focus on recommending emerging items that progress from having no interactions to a few, and eventually to many.

Deep learning models have shown strong ability to capture complex feature interactions and improve click-through rate (CTR) prediction—an essential metric for user engagement with items such as movies, products, or music. However, these models typically rely on large-scale datasets to achieve optimal performance, which is infeasible for emerging items. Due to their large parameter sizes, such models adapt poorly to early phases characterized by sparse interactions, leading to degraded performance and high computational cost.

To alleviate the problem of limited supervision, existing methods can be broadly categorized into three types.
The first type assumes no interactions for emerging items. DropoutNet~\cite{dropoutnet} introduces a dropout mechanism to infer missing data, while later works align representations of new items with those of existing ones~\cite{mtpr,equal,heater,gar,NEURIPS2021_a0443c8c,liu2023enhancing}.
The second type leverages a few interaction records, where few-shot learning~\cite{wang2020generalizing} naturally fits the task. These approaches either adopt gradient-based meta-learning~\cite{melu,mamo,m2eu,edge} or amortization-based strategies~\cite{TaNP,ColdNAS} to adapt models trained on existing items to new ones.
The third type considers items with incrementally increasing interactions, focusing on enhancing item representations~\cite{metaE,mwuf,cvar,gme,aldi,velf}.
Across these approaches, the key principle is to transfer shared patterns from existing items while preserving the uniqueness of emerging ones.

Recognizing the complexity of emerging item recommendation, we leverage the reasoning capabilities and rich knowledge of Large Language Models (LLMs) to learn distinctive embeddings for new items. 
However, achieving this goal is nontrivial. Effective recommendation requires item embeddings that accurately represent new items, align with the existing embedding space, and adapt as new interactions arrive.
To address these challenges, we propose EmerFlow, an LLM-empowered representation learning framework. EmerFlow first enriches the raw features of emerging items through LLM reasoning, then aligns the enhanced representations with the embedding space of the existing model. Finally, it incorporates new interactions through a meta-learning procedure to refine the embeddings. By leveraging limited interactions, EmerFlow learns expressive and distinctive representations for emerging items.
Extensive experiments across domains such as movies and pharmaceuticals demonstrate that EmerFlow consistently outperforms prior methods.

The contributions of this work are summarized as follows:
\begin{itemize}[leftmargin=*]
	\item We propose an LLM-empowered pipeline for emerging item recommendation, where interaction records gradually accumulate from zero.
	\item We combine the general prior knowledge of LLMs with meta-learning to obtain distinctive embeddings for emerging items, preserving their uniqueness while capturing shared patterns during representation learning.
	\item We empirically show that the proposed framework achieves state-of-the-art performance on product recommendation and disease–gene association prediction tasks involving new diseases.
\end{itemize}

\section{Related Works}\label{sec:related_works}

We briefly review four groups of methods pertinent to emerging item recommendation. 

%\noindent
\textbf{Conventional Backbone Models.~} 
Conventional backbone models typically address recommendation tasks without explicitly focusing on underrepresented or emerging items. Their design priorities vary by application domain.
For click-through rate (CTR) prediction of new products, the key challenge lies in modeling complex feature interactions. The field has progressed from first-order models~\cite{lr} to second-order~\cite{fm}, and further to deep architectures that capture high-order interactions~\cite{hofm,wide&deep,deepfm,autoint,cheng2020adaptive,zhu2023final,mao2023finalmlp}.
In contrast, for disease–gene association prediction involving new diseases, the emphasis is on integrating heterogeneous features such as disease phenotypes and gene annotations~\cite{pdgnet,hergepred,KDGene,dada}.
However, these large-parameter models are designed for abundant data and tend to overfit when applied to emerging items with limited interactions. 

%\noindent
\textbf{Approaches for New Items without Interaction Data.~} 
This group of methods tackles the cold-start problem where new items lack any interaction records.
DropoutNet~\cite{dropoutnet} introduces a dropout mechanism to infer missing data, while subsequent studies align representations of new items with those of existing ones~\cite{mtpr,equal,heater,gar,NEURIPS2021_a0443c8c,liu2023enhancing}.
More recently, LLMs have been used to handle new items without interactions~\cite{hou2024large,he2023large}.
Nevertheless, these methods overlook the integration of newly arriving interactions and often require costly retraining to adapt to updates.

%\noindent
\textbf{Approaches for New Items with a Few Interaction Data.~} 
When a small number of interaction records are available, few-shot learning~\cite{wang2020generalizing} provides a natural solution.
These approaches address the $K$-shot setting, where each new item has $K$ labeled instances.
Existing studies adopt either gradient-based meta-learning~\cite{melu,mamo,m2eu,edge} or amortization-based strategies~\cite{TaNP,ColdNAS} to adapt models trained on existing items to new ones.
Recent advances also leverage LLMs for reasoning with minimal data~\cite{huang2024large,sanner2023large}.
However, these methods still struggle to adapt when interaction data are extremely sparse and usually require manual updates to incorporate newly accumulated information.

%\noindent
\textbf{Approaches for New Items with Incremental Interaction Data.~} 
To model the dynamic evolution of emerging items—from zero to few and then many interactions—recent studies focus on improving item representations~\cite{metaE,mwuf,cvar,gme,aldi,velf,wang2024warming}.
Such dynamic adaptation strategies have been applied in multiple domains, including disease–gene association~\cite{disease-gene}, miRNA–disease association~\cite{miRNA-disease}, and new drug recommendation~\cite{edge}.
Building on this foundation, we propose an LLM-empowered representation learning framework that can flexibly accommodate diverse recommendation scenarios.

\section{Problem Formulation}
\label{sec:problem}

In this paper, we address the challenge of recommending emerging items.
We consider a set of items of the same type, where each item $v$ is associated with $N_f$ features $\cF={f_m}_{m=1}^{N_f}$.
For example, a product may include attributes such as item ID, category, and price, while a disease can be characterized by its phenotype and semantic type.
Items with sufficient interaction records are referred to as old items, whereas those with sparse interactions are termed new items.

During training, our goal is to learn a model from a set of recommendation tasks
$\cT^{\text{old}}={\cT_i}{i=1}^{N_t}$ constructed from old items.
The model should generalize quickly to new items that are unseen during training.
Each task $\cT_i$ corresponds to an old item $v_i$ and consists of a training set
$\cS_i = {(v_i,u_j,y{i,j})}{j=1}^{N_s}$ with $N_s$ observed interactions for $v_i$,
and a test set $\cQ_i = {(v_i,u_j,y{i,j})}{j=1}^{N_q}$ with $N_q$ instances for evaluation.
An interaction between an item $v_i$ and a target $u_j$
(e.g., a user in CTR prediction or a gene in disease–gene association)
is labeled as $y{i,j}=1$ for a positive interaction and $y_{i,j}=0$ otherwise.

During testing, we focus on recommending for a new item $v_k$ that initially has no interaction records but gradually accumulates them over time.
To simulate this process, we divide the interactions into multiple incremental phases.
Phase~0 represents the cold-start condition with no interactions.
Each subsequent phase (Phase~I, Phase~II, etc.) introduces an additional set of $K$ new interaction records for $v_k$.
As phases progress, the total number of interactions grows until it reaches a sufficient level.
After each phase, we evaluate the model on a consistent hold-out test set for $v_k$ to measure its performance under progressively richer supervision.

\section{The Proposed \TheName{}}\label{methodology}

In dynamic recommendation systems, the appearance of new items and the gradual accumulation of their interaction records pose a unique challenge.
The effectiveness of recommendation depends on the embedding’s ability to
(i) accurately represent the new item,
(ii) align with the embedding space of existing items so that the backbone model can effectively process it, and
(iii) adapt to new interaction records to achieve rapid performance improvement as data accumulates.

\begin{figure*}[htbp]
	\centering
		\includegraphics[width=1\textwidth]{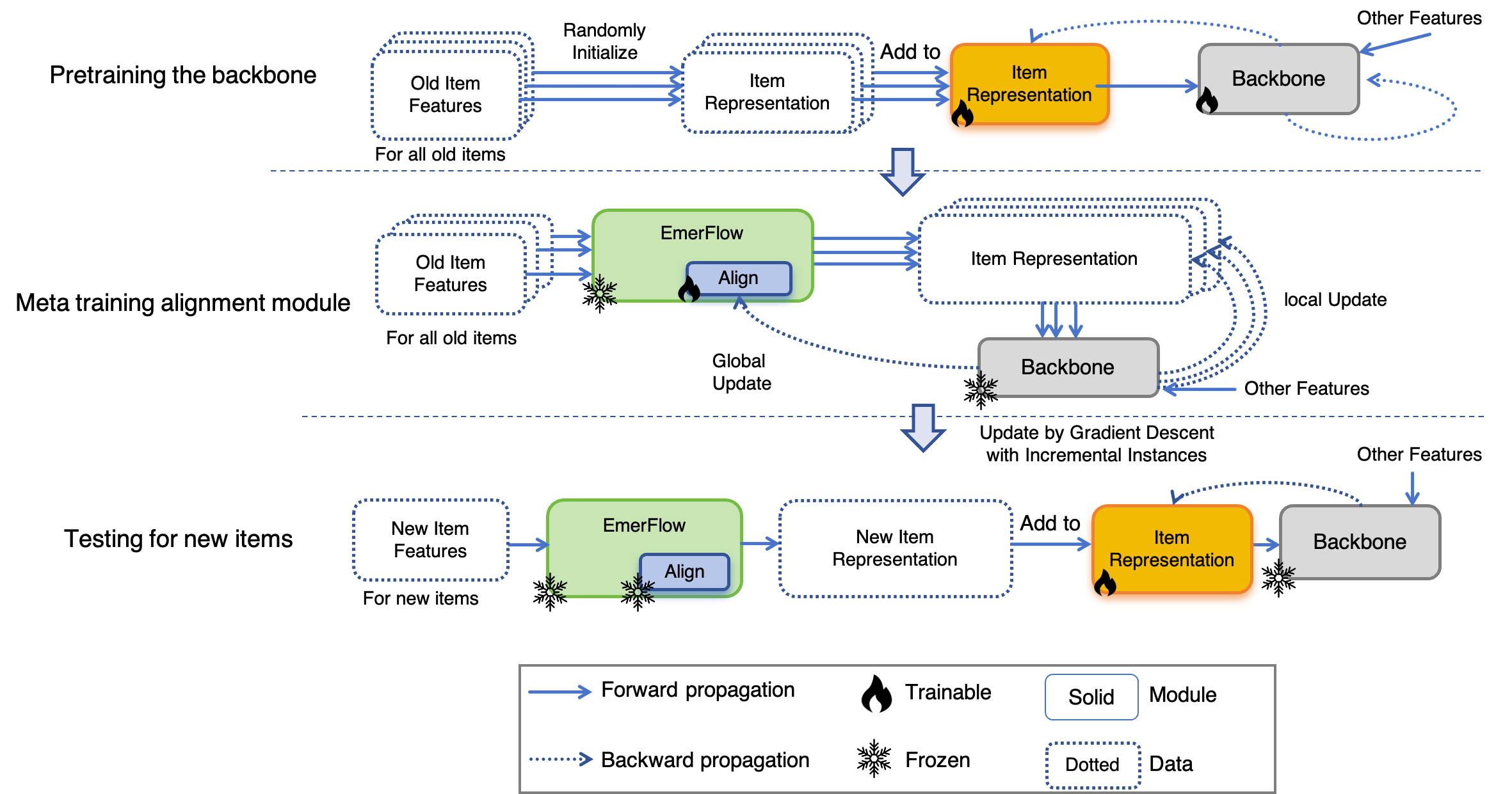}
		%	}
	\caption{
		Illustration of the proposed \TheName{}. 
		For a new item $v_k$, the LLM is prompted to augment its raw features $\cF_k$ with the most relevant auxiliary features $\cA_k$.
		The enriched textual features are then processed by a text feature extractor to obtain the item’s individual representation $\br_k$.
		Finally, an MLP trained with meta learning produces the representative embedding $\be_k$.
		An efficient learning strategy is further designed to minimize training overhead.
	}
	\label{fig:illus}
	\vspace{-10pt}
\end{figure*}

To address these requirements, we propose \TheName{} (Figure~\ref{fig:illus}), a framework designed to satisfy all three properties above.
It consists of three key components:
(i) feature augmentation through LLM reasoning,
(ii) representation alignment, and
(iii) an effective meta-learning strategy.

\subsection{Feature Augmentation}
\label{augmenter}

As discussed in Section~\ref{sec:related_works}, traditional methods transform item features into representations for new items.
However, these features are often limited and may not directly relate to the recommendation task, leading to misaligned representations.

To overcome this limitation, our method leverages the reasoning capability and extensive factual knowledge of LLMs to identify and augment the most relevant features.
We begin by describing the recommendation scenario without interaction records to simulate real-world applications.
We let an LLM reason about the item type and suggest additional features that are absent from the original feature set but likely relevant to potential interactions.
For simplicity, this process is denoted as
\begin{align}\label{eq:aug-feature}
	\cA = \text{LLM}(\text{type}(v)),
\end{align}
where $\text{type}(\cdot)$ returns the type of the item (e.g., movie or disease), and $\text{LLM}(\cdot)$ denotes the language model.
Given the suggested relevant features $\cA$, the LLM is further instructed to provide detailed feature contents for each item $v_i$, resulting in the augmented feature set $\cA_i$.

\begin{figure*}[t]
	\centering
	\subfigure[New products.]{
		\includegraphics[width=0.45\textwidth]{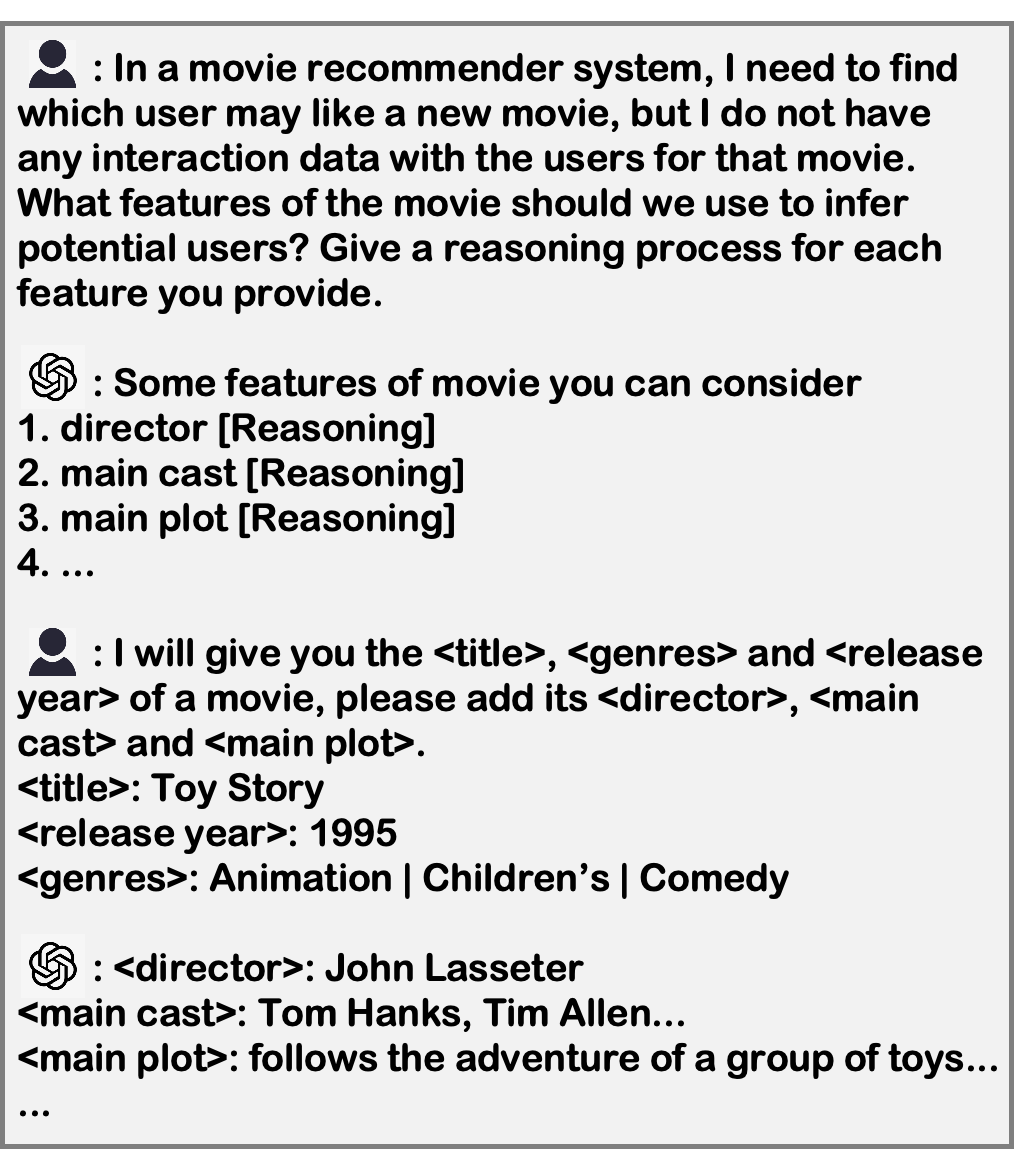}
	}
	~
	\subfigure[New diseases.]{
		\includegraphics[width=0.45\textwidth]{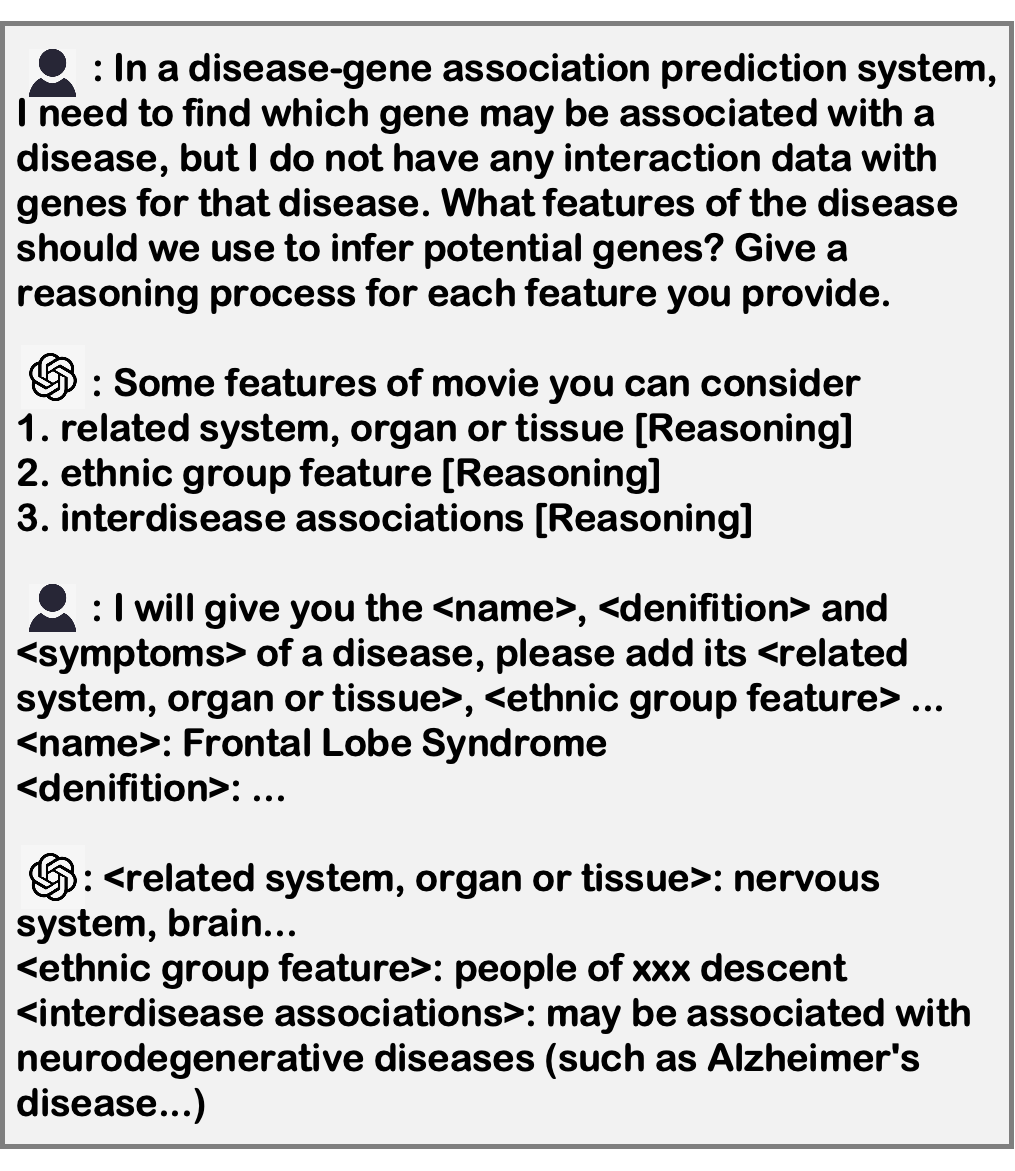}
	}
	\caption{(a) Click-through rate (CTR) prediction of new products: For a movie recognized by the LLM, the model analyzes its context and augments missing features absent from the dataset.
		(b) Disease–gene association prediction of new diseases: For an emerging disease unknown to the LLM, it infers potential relationships, such as linking it to Alzheimer’s disease, based on available features.}
	\label{fig:prompt}
\end{figure*}

Two illustrative examples are shown in Figure~\ref{fig:prompt}.
(i) For new movies, the LLM can reason about the context and enrich missing features not present in the dataset.
(ii) For emerging diseases, the LLM can infer relevant biomedical features that require domain knowledge, such as linking an unfamiliar disease to Alzheimer’s disease.

Since the augmented features $\cA_i$ generated by the LLM are textual, whereas the original features $\cF_i$ are typically tabular, we unify the processing of both types through text serialization using a predefined template, following~\cite{tabLLM}.
Each feature $f_m$ is serialized as
\begin{align}
	\text{serialize}(f_m) = \text{<name($f_m$)>: content($f_m$)},
\end{align}
where $\text{name}(\cdot)$ returns the descriptor of the feature (e.g., “Year”), and $\text{content}(\cdot)$ returns its specific value (e.g., “1992”).
For instance, a movie released in 1992 has its feature “Release Year” serialized as
\text{<Release Year>: 1992}.
Concatenating all serialized features forms a comprehensive string $s_i$, which is then fed into a text feature encoder:
\begin{align}\label{eq:encoder}
	\br_i = \text{encoder}(s_i).
\end{align}

\subsection{Representation Alignment}\label{meta-learner}

The LLM-empowered item representation $\br_i$ encodes extensive knowledge captured by the language model.
However, it cannot be directly used by conventional backbone models.
The backbone models are trained on numerous old items with their observed interactions, while the LLM is pretrained on massive text corpora.
As a result, their embedding spaces differ, and features in corresponding dimensions carry distinct semantic meanings.

To bridge this gap, we introduce an alignment module to map the LLM-based representations into the embedding space of the backbone models.
Directly retraining or fine-tuning both models would be computationally expensive; therefore, we train only a lightweight alignment network while keeping both the backbone and the LLM fixed.
This strategy produces an item representation that is compatible with the backbone model for subsequent recommendation.

Formally, let the alignment module be denoted as $\text{align}(\cdot)$.
The updated item representation $\be_i$ is computed as
\begin{equation}\label{eq:align}
	\be_i = \text{align}(\br_i).
\end{equation}

\subsection{Learning and Inference} 
\label{sec:alg}

In \TheName{}, the backbone models are pretrained on old items with abundant interaction records, as discussed in Section~\ref{sec:related_works}, while the LLM remains fixed to reduce computational cost.
This design allows the use of API-based LLMs and existing pretrained textual feature encoders.
Therefore, we only consistently optimize the alignment module $\text{align}(\cdot)$ in \eqref{eq:align} across all phases,
and update the item representation $\be_i$ using the incremental interaction records of $v_i$.

Let $\btheta$ denote the trainable parameters of $\text{align}(\cdot)$.
We optimize \TheName{} with respect to the following meta-objective over $N_t$ tasks sampled from $\cT^{\text{old}}$:
\begin{equation}\label{eq:loss-meta}
	\sum\nolimits_{i=1}^{N_t}
	\mL_{\cS_i}(\btheta,\be_i)
	+
	\alpha
	\mL_{\cQ_i}(\btheta,\be_i'),
\end{equation}
where $\alpha$ balances the two loss terms.
Specifically, $\mL_{\cS_i}(\btheta,\be_i)$ represents the Phase-0 performance~\cite{metaE}, calculated as
\begin{align}\label{eq:loss-s}
	\frac{1}{|\cS_i|}\sum_{(v_i,u_j,y_{ij})\in\cS_i}
	\text{BCE}(y_{ij},\text{backbone}(\be_i,\cF_i,u_j)),
\end{align}
where
$\text{BCE}(y,\hat{y})=-y\log(\hat{y})-(1-y)\log(1-\hat{y})$,
and $\text{backbone}(\cdot)$ denotes a conventional recommendation model introduced in Section~\ref{sec:related_works}.
$\mL_{\cQ_i}(\btheta,\be_i')$ measures the loss after updating $\be_i$ using a few new item instances from $\cS_i$:
\begin{align}\label{eq:loss-q}
	\frac{1}{|\cQ_i|}\sum_{(v_i,u_j,y_{ij})\in\cQ_i}
	\text{BCE}(y_{ij},\text{backbone}(\be_i',\cF_i,u_j)),
\end{align}
where $\be_i'$ is obtained by taking gradient descent w.r.t. \eqref{eq:loss-s}.

Algorithm~\ref{alg:training} summarizes the training procedure.
By learning from tasks in $\cT^{\text{old}}$, the optimized parameters $\btheta^*$ capture transferable knowledge across item types.

\begin{algorithm}[H]
	\caption{Training procedure of \TheName{}.}
	\label{alg:training}
	\begin{algorithmic}[1]
		\STATE Randomly initialize $\btheta$;
		\WHILE{not converged}
		\STATE Sample a batch of tasks $\cT_i \sim \cT^{\text{old}}$;
		\FORALL{$\cT_i$}
		\STATE Obtain the initial embedding $\be_i$ for item $v_i$;
		\STATE Compute $\mL_{\cS_i}(\btheta,\be_i)$ by \eqref{eq:loss-s};
		\STATE Update $\be_i$ to $\be_i'$ via gradient descent on $\mL_{\cS_i}(\btheta,\be_i)$;
		\STATE Compute $\mL_{\cQ_i}(\btheta,\be_i')$ by \eqref{eq:loss-q};
		\ENDFOR
		\STATE Update $\btheta$ using the meta-objective in \eqref{eq:loss-meta};
		\ENDWHILE
	\end{algorithmic}
\end{algorithm}

During testing, we consider a task $\cT_k$ corresponding to a new item $v_k$.
When $v_k$ has no interaction records (Phase-0), we directly evaluate its performance on the hold-out test set.
In subsequent phases, as a few interaction instances become available, we update the item representation from $\be_i$ to $\be_i'$ to incorporate the new supervision and re-evaluate performance on the same test set.

\section{Experiments}

We conduct experiments on two representative tasks:
(i) predicting click-through rates (CTR) for newly emerging products, and
(ii) predicting disease–gene associations for emerging diseases.
These experiments are designed to evaluate the effectiveness of \TheName{} in handling different types of emerging item recommendation scenarios.

\subsection{Experimental Settings}
\label{sec:settings}

\subsubsection{Datasets}
\label{app:data}
We evaluate \TheName{} on two representative datasets.

\textbf{CTR prediction for new products.}
We use the \emph{MovieLens}~\cite{movielens} dataset, where each item corresponds to a movie.
The dataset contains 1 million user–movie ratings (from 1 to 5).
Following~\cite{metaE,cvar}, ratings below 4 are binarized as negative (0), and the rest as positive (1).
Each movie is associated with attributes such as movie ID, title, release year, and genres, while each user has user ID, gender, occupation, and zip code.

\textbf{Disease–gene association prediction for new diseases.}
We formulate this task as a recommendation problem where each disease is treated as an item, and the goal is to predict whether a gene is associated with the disease.
We use the \emph{DisGeNet}~\cite{disgenet} dataset, which contains approximately 1.1 million disease–gene associations with confidence scores representing reliability levels.
These associations involve 30,170 diseases and 20,671 genes.
Each gene includes attributes such as gene ID and protein class, while each disease includes disease ID and semantic type.
Associations with scores above 0.01 are regarded as positive samples, and negative samples are randomly drawn from unobserved disease–gene pairs.

\textbf{Data split and phases.}
Following~\cite{metaE,cvar}, items are divided into two groups according to their interaction frequency:
(i) \textbf{Old items}—items with more than $N$ records;
(ii) \textbf{New items}—items with no more than $N$ but at least $3K$ records.
We set $N=200$ and $K=20$ for both datasets.
To simulate realistic scenarios of emerging items, we sort the interactions of each item by timestamp (for \emph{MovieLens}) or by the first referenced year (for \emph{DisGeNet}).
The interaction records corresponding to old items are used as training set for pretraining. Then for each new item, its corresponding records are split into 4 sets, where the first 3 sets contain $K$ records and the last set contain the remaining records. These 4 sets of records are used to test for Phase-0, Phase-I, Phase-II and Phase-III respectively. After test during each phase, the set for test in this phase is used to optimize the item representation and then the next set is used to test for next phase.

\textbf{Evaluation metrics.}
We report (i) the Area Under the ROC Curve (AUC), which measures the ranking separability between positive and negative samples, and (ii) the F1 score, the harmonic mean of precision and recall.
Both metrics range from 0 (worst) to 1 (best).

\subsubsection{Baselines}
\label{app:baseline}

For each application, we compare \TheName{} with a broad range of baselines, including conventional backbone models and approaches specifically designed for new items.

\textbf{Conventional backbone models.}
These models are widely used in recommendation systems but do not explicitly address underrepresented items.
For CTR prediction of new products, we include
DeepFM~\cite{deepfm}, Wide\&Deep~\cite{wide&deep}, AutoInt~\cite{autoint}, AFN~\cite{cheng2020adaptive}, FinalMLP~\cite{mao2023finalmlp}, and FINAL~\cite{zhu2023final}.
For disease–gene association prediction of new diseases, we evaluate
PDGNet~\cite{pdgnet}, HerGePred~\cite{hergepred}, DADA~\cite{dada}, and KDGene~\cite{KDGene}.

\textbf{Methods for emerging items.}
We further compare with three groups of approaches focusing on new items:
(i) \emph{Methods without interaction data:} DropoutNet~\cite{dropoutnet} and ALDI~\cite{aldi};
(ii) \emph{Methods with a few interaction records:} MAML~\cite{maml}, ProtoNet~\cite{protoNet}, EGNN~\cite{EGNN}, and ColdNAS~\cite{ColdNAS};
(iii) \emph{Methods with incremental interaction data:} MetaE~\cite{metaE}, CVAR~\cite{cvar}, and MWUF~\cite{mwuf}.

For all baseline methods, we use publicly available implementations released by the original authors.
Hyperparameters are tuned via grid search on the validation set for fair comparison.
We use DeepFM~\cite{deepfm} and PDGNet~\cite{pdgnet} as the backbone models for the two applications, respectively.
In \TheName{}, ChatGPT-3.5 serves as the LLM for feature augmentation, and LLaMA-2-7B~\cite{llama} is adopted as the text encoder $\text{encoder}(\cdot)$.
The pretraining corpus of ChatGPT-3.5 used for augmenting features may contain classical datasets and disease–gene data, but this LLM is used for augmenting features in the form of text. We do not find that it provides associations in augmented features. For pretraining corpus of open-source LLaMA-2-7B used for feature encoding, it dose not include structured biomedical databases such as DisGeNet nor curated rating datasets such as MovieLens.

\begin{table*}[htbp]
	\centering
	\caption{Test performance obtained on \emph{MovieLens}. 
		The best results are bolded, the second-best results are underlined. 
	}
		\begin{tabular}{c|cc|cc|cc|cc}
			\hline
			\multirow{2}{*}{\emph{MovieLens}} & \multicolumn{2}{c|}{Phase-0} & \multicolumn{2}{c|}{Phase-$\text{\rnum{1}}$} & \multicolumn{2}{c|}{Phase-$\text{\rnum{2}}$} & \multicolumn{2}{c}{Phase-$\text{\rnum{3}}$} \\	
			& AUC(\%) & F1(\%) & AUC(\%) & F1(\%) & AUC(\%) & F1(\%) & AUC(\%) & F1(\%)\\	\hline
			DeepFM & \ms{72.54}{0.15} & \ms{62.61}{0.22} & \ms{75.21}{0.27} & \ms{64.12}{0.01} & \ms{75.84}{0.26} & \ms{65.30}{0.15} & \ms{76.98}{0.19} & \ms{66.05}{0.14} \\
			Wide\&Deep & \ms{71.27}{0.29} & \ms{59.36}{0.31} & \ms{72.84}{0.29} & \ms{61.57}{0.21} & \ms{73.99}{0.25} & \ms{62.76}{0.24} & \ms{74.97}{0.22} & \ms{63.64}{0.27} \\
			AutoInt & \ms{70.20}{0.24} & \ms{60.66}{0.14} & \ms{72.88}{0.31} & \ms{62.75}{0.36} & \ms{74.96}{0.33} & \ms{64.32}{0.42} & \ms{76.57}{0.34} & \ms{65.66}{0.36} \\
			AFN & \ms{71.45}{0.42} & \ms{61.67}{0.37} & \ms{73.57}{0.08} & \ms{63.61}{0.09} & \ms{75.49}{0.24} & \ms{65.24}{0.16} & \ms{76.94}{0.35} & \ms{66.48}{0.28} \\
			FinalMLP& \ms{69.51}{0.06} & \ms{60.59}{0.17} & \ms{78.48}{0.12} & \ms{67.34}{0.10} & \ms{78.47}{0.16} & \ms{67.27}{0.07} & \ms{79.07}{0.17} & \ms{68.00}{0.10}\\
			FINAL& \ms{71.64}{0.15} & \ms{61.72}{0.17} & \ms{77.87}{0.13} & \ms{66.99}{0.10} & \ms{77.94}{0.10} & \ms{66.93}{0.14} & \ms{78.29}{0.09} & \ms{67.42}{0.10}\\
			\hline
			DropoutNet & \ms{73.30}{0.02} & \ms{63.23}{0.02} &-& - & - &-& -&- 
			\\
			ALDI & \ms{65.53}{0.13} & \ms{57.47}{0.23} &-& - & - &-& -&- \\
			\hline
			MAML & -&- & \ms{77.24}{0.06} & \ms{66.53}{0.11} & \ms{79.29}{0.10} & \ms{68.11}{0.05} & \ms{80.23}{0.03} & \ms{68.16}{0.06} \\
			ProtoNet & - & - & \ms{78.03}{0.23} & \ms{65.94}{0.15} & \ms{79.56}{0.11} & \ms{68.11}{0.13} & \ms{80.01}{0.08} & \ms{67.98}{0.07} \\
			EGNN & - & - & \ms{78.87}{0.34} & \ms{67.23}{0.19} & \ms{79.72}{0.21} & \ms{68.24}{0.13} & \ms{80.33}{0.14} & \ms{68.27}{0.11} \\
			ColdNAS & - & - & \ms{77.45}{0.03} & \ms{67.01}{0.03} & \ms{77.88}{0.12} & \ms{67.25}{0.21} & \ms{78.06}{0.09} & \ms{67.31}{0.11} \\\hline
			MetaE & \ms{72.10}{0.70} & \ms{63.06}{0.30} & \ms{77.57}{0.25} & \ms{67.06}{0.15} & \ms{79.87}{0.09} & \bms{68.74}{0.12} & \bms{80.69}{0.04} & \bms{69.31}{0.12} \\
			CVAR & \sbms{73.69}{0.21} & \sbms{63.25}{0.12} & \sbms{78.99}{0.10} & \sbms{67.75}{0.26} & \ms{80.15}{0.06} & \ms{68.57}{0.12} & \ms{80.39}{0.04} & \ms{68.73}{0.14} \\
			MWUF & \ms{73.19}{0.66} & \ms{62.61}{0.74} & \ms{78.88}{0.11} & \ms{67.34}{0.22} & \sbms{80.26}{0.08} & \ms{68.40}{0.13} & \ms{80.57}{0.05} & \ms{68.66}{0.10} \\
			\TheName{} & \bms{78.17}{0.54} & \bms{66.17}{0.23} & \bms{79.96}{0.17} & \bms{68.31}{0.10} & \bms{80.38}{0.07} & \sbms{68.62}{0.12} & \sbms{80.60}{0.04} & \sbms{68.90}{0.13} \\
			\hline
		\end{tabular}
	\label{tab:result_movielens_full}
\end{table*}

\subsection{Performance Comparison}\label{performance}

\textbf{CTR prediction for new products.}
Table~\ref{tab:result_movielens_full} presents the results on \emph{MovieLens}.
Among the baseline groups, the fourth group achieves the best overall performance, as it addresses both Phase-0 and the subsequent accumulation of interaction records.
General recommendation models perform worse because they are not designed for emerging items.
The second group can handle Phase-0 better than the backbone alone but still lags behind \TheName{}, since it ignores incremental interactions.
The third group focuses on few-shot scenarios but cannot handle Phase-0 effectively. 
In comparison, \TheName{} achieves the highest average performance across all phases.
This improvement arises from the rich semantic information encoded in the LLM-empowered item representation, which is well aligned with the embedding space of old items.
Although the margin decreases in Phases~II and~III, this is mainly due to the limited capacity of the backbone model, as confirmed in later experiments.

\textbf{Disease–gene association prediction for new diseases.}
Table~\ref{tab:result_disgenet_full} reports the results on \emph{DisGeNet}.
Notably, the backbone model PDGNet already performs strongly, as it effectively integrates multiple types of biological interactions, including disease–gene, disease–phenotype, and Gene Ontology (GO) associations.
Nevertheless, \TheName{} consistently outperforms all baselines and achieves the best results across all four phases, demonstrating its strong generalization and adaptability to emerging diseases.

\begin{table*}[h]
	\centering
	\caption{Test performance obtained on \emph{DisGeNet}. 
	}
		\begin{tabular}{c|cc|cc|cc|cc}
			\hline
			\multirow{2}{*}{} & \multicolumn{2}{c|}{Phase-0} & \multicolumn{2}{c|}{Phase-$\text{\rnum{1}}$} & \multicolumn{2}{c|}{Phase-$\text{\rnum{2}}$} & \multicolumn{2}{c}{Phase-$\text{\rnum{3}}$}\\	& AUC(\%) & F1(\%) & AUC(\%) & F1(\%) & AUC(\%) & F1(\%) & AUC(\%) & F1(\%) \\ \hline
			PDGNet & \ms{79.33}{0.02} & \ms{72.07}{0.01} & \ms{80.94}{0.09} & \ms{73.78}{0.19} & \ms{82.26}{0.10} & \ms{74.96}{0.21} & \ms{82.82}{0.11} & \ms{75.47}{0.29}  \\
			KDGene & \ms{76.98}{0.02} & \ms{70.68}{0.00} & \ms{75.41}{0.05} & \ms{69.78}{0.07} & \ms{77.09}{0.05} & \ms{71.66}{0.31} & \ms{77.80}{0.07} & \ms{72.17}{0.13}  \\
			DADA & \ms{53.31}{0.01} & \ms{49.54}{0.03} & \ms{72.88}{0.12} & \ms{65.47}{0.02} & \ms{73.84}{0.18} & \ms{66.39}{0.16} & \ms{74.13}{0.02} & \ms{66.95}{0.39}  \\
			HerGePred & \ms{73.82}{0.00} & \ms{68.95}{0.01} & \ms{75.01}{0.02} & \ms{69.10}{0.01} & \ms{76.35}{0.15} & \ms{70.48}{0.11} & \ms{77.18}{0.06} & \ms{70.98}{0.08}  \\
			\hline
			DropoutNet & \ms{79.50}{0.02} & \ms{72.32}{0.01} &  - & - &-& -&-&-  \\
			ALDI& \ms{77.34}{0.03} & \ms{69.53}{0.02} & - & - &-& -&-&- \\
			\hline
			MAML &-& -& \ms{76.15}{0.38} & \ms{69.53}{0.25} & \ms{77.75}{0.32} & \ms{71.26}{0.17} & \ms{79.03}{0.24} & 
			\ms{72.47}{0.20}
			\\
			ProtoNet & - & - & \ms{79.33}{0.17} & \ms{72.87}{0.23} & \ms{81.53}{0.09} & \ms{73.32}{0.14} & \ms{82.39}{0.05} & \ms{75.03}{0.03}  \\
			EGNN & - & - & \ms{79.43}{0.12} & \ms{72.90}{0.15} & \ms{81.69}{0.08} & \ms{73.64}{0.09} & \ms{82.42}{0.03} & \ms{75.05}{0.02}  \\
			ColdNAS& - & - & \ms{79.05}{0.07} & \ms{72.44}{0.05} & \ms{80.76}{0.10} & \ms{72.87}{0.08} & \ms{81.92}{0.04} & \ms{74.83}{0.06} \\
			\hline
			MetaE & \sbms{79.77}{0.19} & \sbms{72.44}{0.85} & \ms{80.93}{0.15} & \sbms{74.01}{0.38} & \ms{82.26}{0.18} & \sbms{75.12}{0.49} & \sbms{82.90}{0.11} & \sbms{75.62}{0.39} \\
			CVAR & \ms{77.37}{0.45} & \ms{69.84}{0.31} & \ms{79.14}{0.50} & \ms{72.22}{0.61} & \ms{80.01}{0.07} & \ms{72.94}{0.10} & \ms{80.61}{0.34} & \ms{73.52}{0.47} \\
			MWUF & \ms{79.26}{0.09} & \ms{70.81}{0.33} & \sbms{80.96}{0.01} & \ms{73.30}{0.38} & \sbms{82.29}{0.03} & \ms{74.81}{0.03} & \ms{82.88}{0.03} & \ms{75.46}{0.17} \\
			\TheName{} & \bms{81.04}{0.03} & \bms{73.38}{0.07} & \bms{81.84}{0.02} & \bms{74.73}{0.08} & \bms{82.83}{0.02} & \bms{75.60}{0.03} & \bms{83.21}{0.02} & \bms{75.74}{0.03} \\
			\hline
		\end{tabular}
		%	}
	\label{tab:result_disgenet_full}
\end{table*}

\begin{figure}[htbp]
	\vspace{-10pt}
	\centering
	\subfigure[\emph{MovieLens}.]{
		\includegraphics[width=0.23\textwidth]{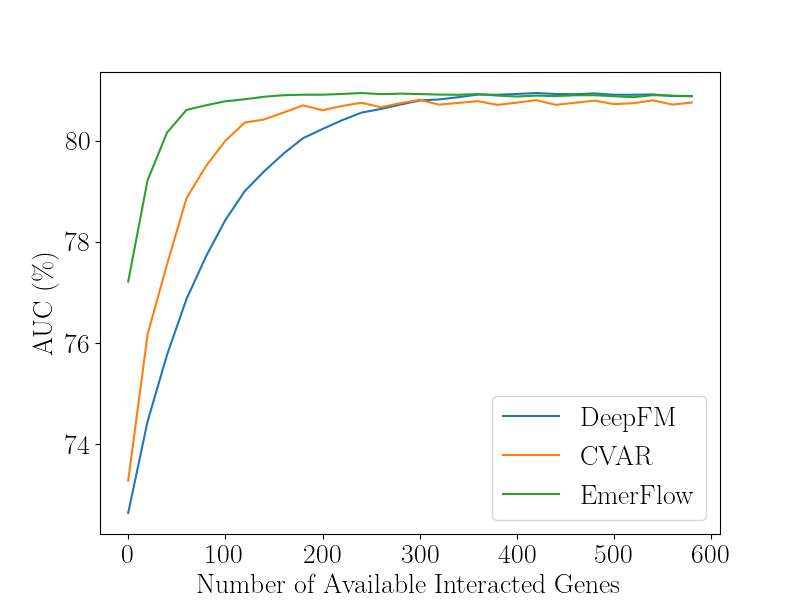}
	}
	\subfigure[\emph{DisGeNet}.]{
		\includegraphics[width=0.23\textwidth]{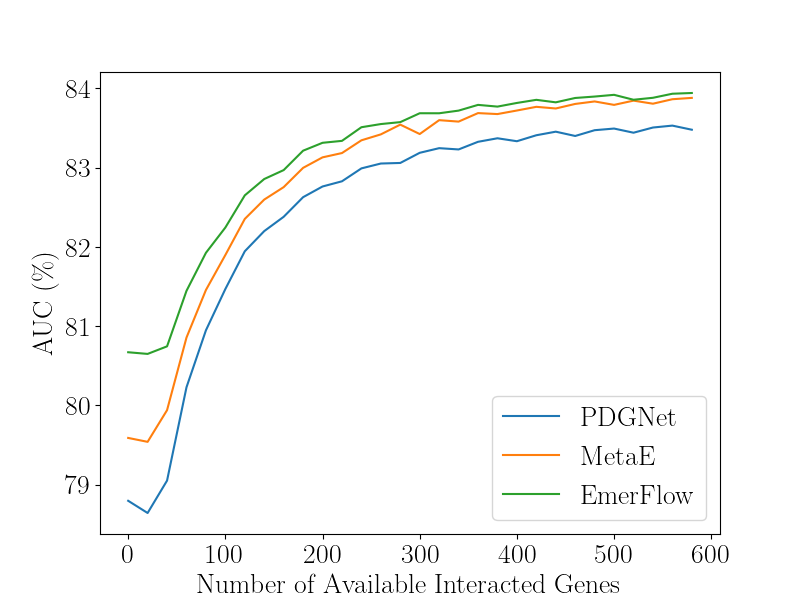}
	}
	\vspace{-10pt}
	\caption{Comparing \TheName{} with the second-best baselines  and the backbone models with varying numbers of training samples. 
		}
		
	\label{fig:abundance-AUC}
\end{figure}

\textbf{Accumulating Sufficient Training Data.~}  
We further analyze the process of gradually increasing the number of interaction records for a new item—from none to a sufficient amount.
To simulate this, we reserve part of the original test set of new items (resulting in a smaller test set), use the reserved samples to augment the training data, and evaluate performance on the reduced test set.
We compare \TheName{} with the second-best baselines—CVAR for \emph{MovieLens} and MAML for \emph{DisGeNet}—as well as with the backbone models.
Figure~\ref{fig:abundance-AUC} shows the testing AUC (\%) as the number of training samples increases.
When training data are limited, \TheName{} consistently achieves the best performance.
Given more data, its performance gradually converges to that of the backbone models and other strong baselines.
After convergence, \TheName{} performs comparably to, or slightly better than, the conventional backbones.
This convergence level indicates that the performance gap narrows in Phases~II and~III (as also observed in Table~\ref{tab:result_movielens_full}), since the results approach the upper limit of the backbone model.
\begin{table*}[h]
	\centering
	\caption{Comparing with TabLLM on \emph{MovieLens}. 
	}
	\begin{tabular}{c|cc|cc|cc|cc}
		\hline
		\multirow{2}{*}{} & \multicolumn{2}{c|}{Phase-0} & \multicolumn{2}{c|}{Phase-$\text{\rnum{1}}$} & \multicolumn{2}{c|}{Phase-$\text{\rnum{2}}$} & \multicolumn{2}{c}{Phase-$\text{\rnum{3}}$} \\	& AUC(\%) & F1(\%) & AUC(\%) & F1(\%) & AUC(\%) & F1(\%) & AUC(\%) & F1(\%)\\ \hline
		TabLLM & \ms{56.24}{0.14} & \ms{53.25}{0.08} & \ms{56.76}{0.25} & \ms{53.78}{0.17} & \ms{57.18}{0.13} & \ms{54.02}{0.10} & \ms{57.16}{0.16} & \ms{54.04}{0.11} \\
		\TheName{} & \bms{78.17}{0.54} & \bms{66.17}{0.23} & \bms{79.96}{0.17} & \bms{68.31}{0.10} & \bms{80.38}{0.07} & \bms{68.62}{0.12} & \bms{80.60}{0.04} & \bms{68.90}{0.13} \\
		\hline
	\end{tabular}
	\label{tab:LLM}
\end{table*}
\begin{table*}[h]
	\centering
	\caption{Test performance of different transformer-based models as $\text{encoder}(\cdot)$
	}
	\resizebox{1\textwidth}{!}{
		\begin{tabular}{c|cc|cc|cc|cc}
			\hline
			\multirow{2}{*}{\emph{MovieLens}} & \multicolumn{2}{c|}{Phase-0} & \multicolumn{2}{c|}{Phase-$\text{\rnum{1}}$} & \multicolumn{2}{c|}{Phase-$\text{\rnum{2}}$} & \multicolumn{2}{c}{Phase-$\text{\rnum{3}}$} \\	
			& AUC(\%) & F1(\%) & AUC(\%) & F1(\%) & AUC(\%) & F1(\%) & AUC(\%) & F1(\%)\\	\hline
			BERT & \ms{73.72}{0.17} & \ms{63.50}{0.08} & \ms{77.49}{0.20} & \ms{66.47}{0.07} & \ms{79.30}{0.26} & \ms{67.97}{0.12} & \ms{80.09}{0.28} & \ms{68.57}{0.13} \\
			BERT-Large & \ms{74.54}{0.14} & \ms{64.15}{0.11} & \ms{77.78}{0.15} & \ms{66.68}{0.11} & \ms{79.41}{0.06} & \ms{67.98}{0.11} & \ms{80.13}{0.21} & \ms{68.51}{0.14} \\
			LLaMa2-7B & \sbms{78.17}{0.54} & \sbms{66.17}{0.23} & \sbms{79.96}{0.17} & \sbms{68.31}{0.10} & \sbms{80.38}{0.07} & \sbms{68.62}{0.12} & \sbms{80.60}{0.04} & \sbms{68.90}{0.13} \\
			LLaMa2-7B-Chat & \ms{75.80}{0.15} & \ms{65.27}{0.16} & \ms{78.41}{0.16} & \ms{67.25}{0.19} & \ms{79.75}{0.24} & \ms{68.41}{0.14} & \ms{80.35}{0.18} & \ms{68.84}{0.06} \\
			LLaMa2-13B & \bms{78.35}{0.43} & \bms{66.42}{0.20} & \bms{80.18}{0.20} & \bms{68.89}{0.14} & \bms{80.97}{0.10} & \bms{69.01}{0.09} & \bms{81.02}{0.11} & \bms{69.14}{0.07} \\
			LLaMa2-13B-Chat & \ms{77.02}{0.15} & \ms{65.84}{0.16} & \ms{79.11}{0.16} & \ms{67.74}{0.19} & \ms{80.03}{0.24} & \ms{68.50}{0.14} & \ms{80.44}{0.18} & \sbms{68.91}{0.06} \\
			\hline
		\end{tabular}
	}
	\label{tab:result_different_Ext}
\end{table*}

\subsection{Ablation Study}
We compare \TheName{} with the following variants:
(i) \textbf{w/o LLM}: This variant removes the LLM and directly uses $\text{encoder}(\cdot)$ to encode the serialized raw item features;
(ii) \textbf{w/o $\cF_i$}: $\text{encoder}(\cdot)$ encodes only the augmented features $\cA_i$, excluding the original features $\cF_i$; and
(iii) \textbf{w/o $\cL_{\cQ_i}$}: This variant omits $\cL_{\cQ_i}$ in \eqref{eq:loss-meta}.

\begin{figure}[h]
	\centering
	\centering
	\includegraphics[width=0.45\textwidth]{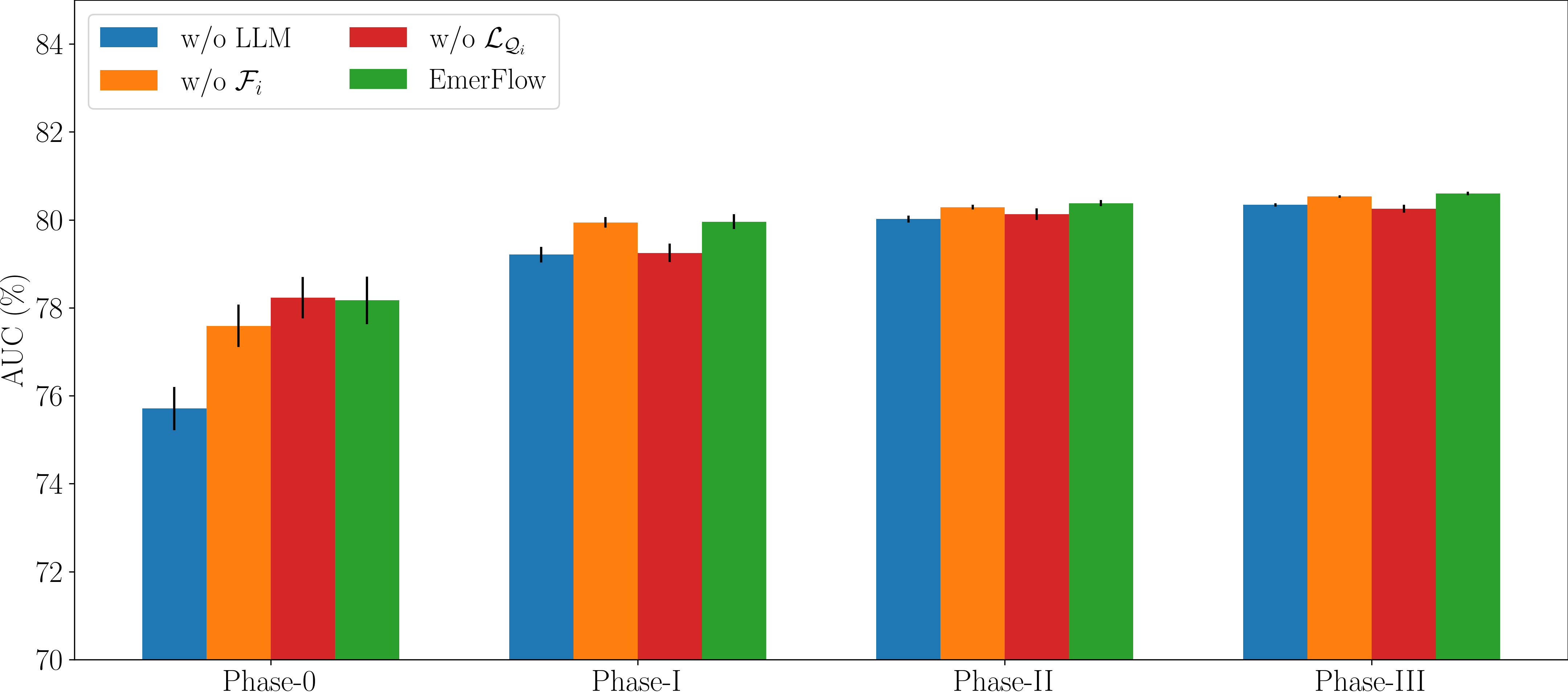}
	\caption{Ablation study on \emph{MovieLens}.}
	\vspace{-20pt}
	\label{fig:ablation-backbone}
\end{figure}
Figure~\ref{fig:ablation-backbone} presents the results.
\TheName{} outperforms w/o LLM, particularly in Phase-0.
This improvement comes from the LLM’s ability to supplement external prior knowledge absent from the dataset, producing semantically richer item representations.
We also investigate the correctness of augmented feature with randomly sampled 100 movies from MovieLens and 100 diseases from DisGeNet. We find that the features generated by LLM is 98\% correct for sampled movies and 96\% correct for sampled diseases.
These results show that almost all features augmented by LLM are correct.
Interestingly, w/o $\cF_i$ achieves performance similar to \TheName{} in both domains, likely because the augmented features already embed most of the information contained in the original features.
w/o $\cL_{\cQ_i}$ performs slightly better than \TheName{} in Phase-0 but shows slower improvement in later phases.
This behavior aligns with previous findings~\cite{maml}, indicating that certain parameter patterns are more sensitive to task-specific optimization.
Without $\cL_{\cQ_i}$, the model does not fine-tune item representations within each task, so optimization efficiency is not explicitly encouraged.
However, since it only aligns LLM-generated representations with those of old items via $\cL_{\cS_i}$, this variant naturally performs well in Phase-0, where no interaction data are available.
\subsection{Model Analysis}

\textbf{Comparison with an LLM-Based Method.}
We compare our proposed framework with a recent LLM-based baseline, TabLLM~\cite{tabLLM}, which is specifically designed for tabular data but relies solely on in-context learning.  
As shown in Table~\ref{tab:LLM}, TabLLM struggles with CTR prediction, indicating that simply adopting an LLM is insufficient for emerging item recommendation.  
In contrast, \TheName{} consistently achieves higher AUC and F1 scores across all phases, validating the necessity of our workflow that integrates LLM reasoning, representation alignment, and meta-learning.

\textbf{Influence of Different Encoders.}
In the main experiments, we use LLaMA2-7B as the $\text{encoder}(\cdot)$.  
To examine the effect of different encoders, we replace it with various transformer-based models, including BERT, BERT-Large, LLaMA2-7B-Chat, LLaMA2-13B, and LLaMA2-13B-Chat.  
Results on \emph{MovieLens} (Table~\ref{tab:result_different_Ext}) show that performance improves as the parameter scale of $\text{encoder}(\cdot)$ increases.  
This observation highlights that strong general text encoders are more suitable for our setting.

\begin{figure*}[htbp]
	\centering
	\subfigure[DeepFM.]{
		\includegraphics[width=0.23\textwidth]{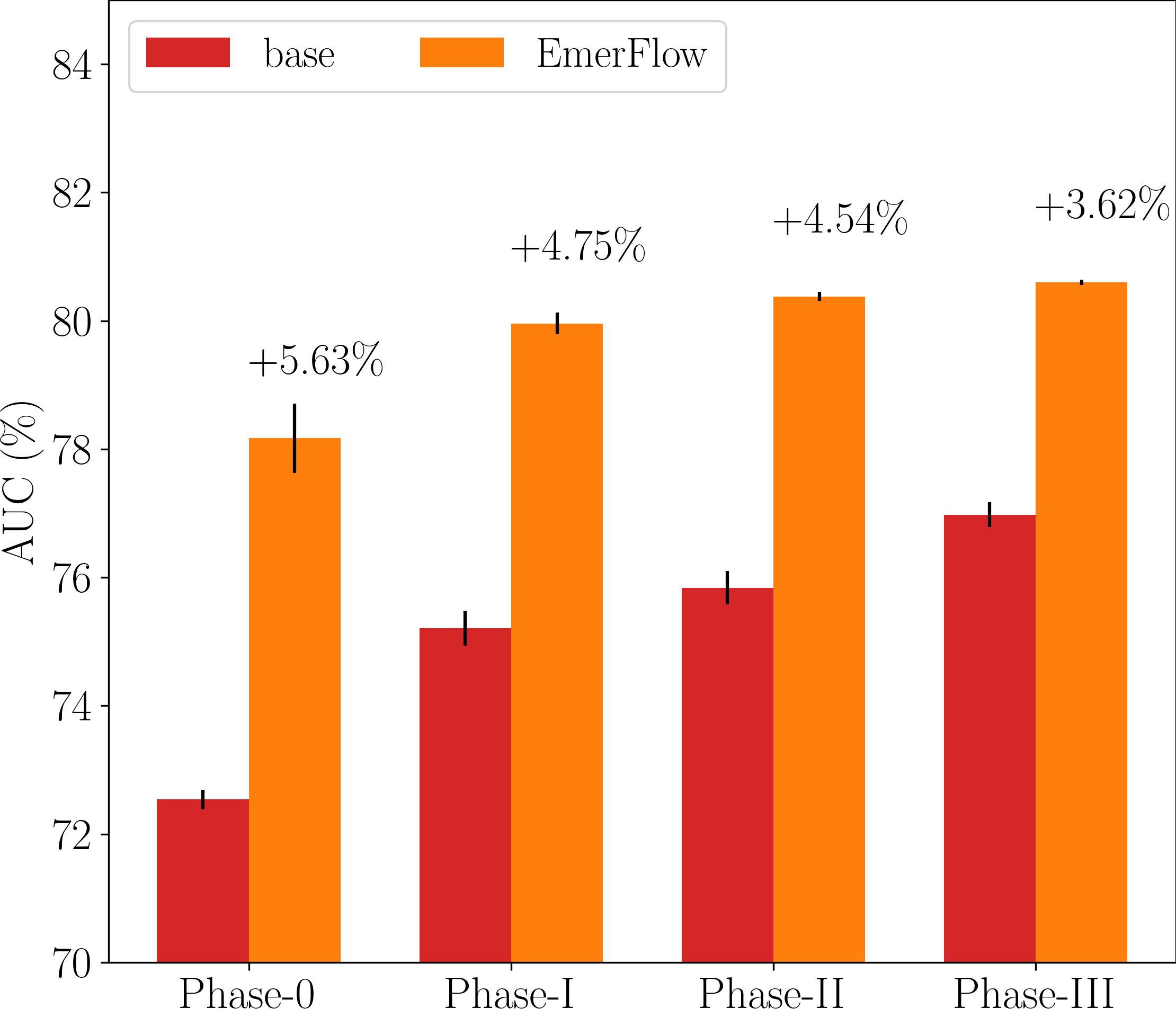}
	}
	\subfigure[Wide\&Deep.]{
		\includegraphics[width=0.23\textwidth]{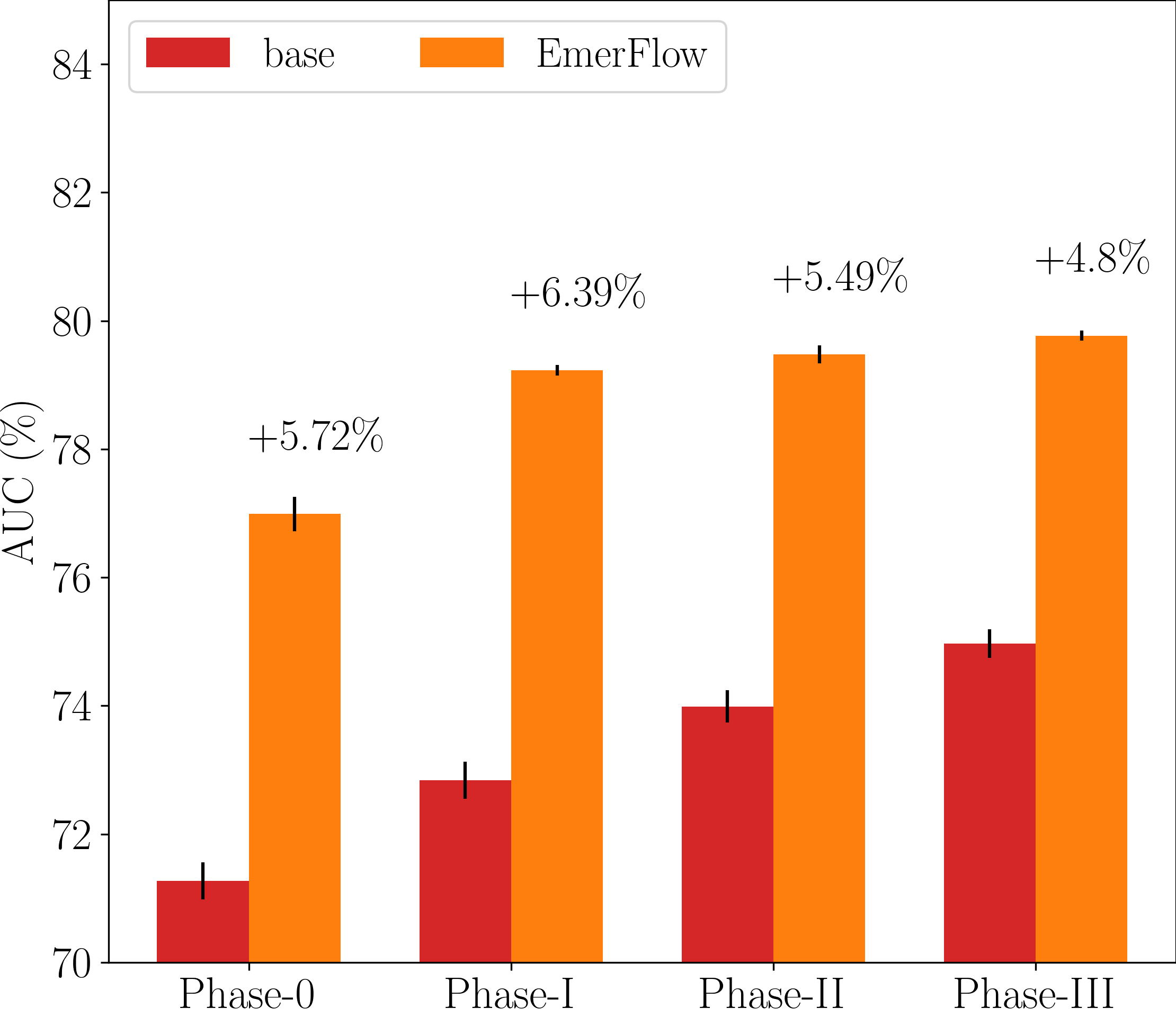}
	}
	\subfigure[AFN.]{
		\includegraphics[width=0.23\textwidth]{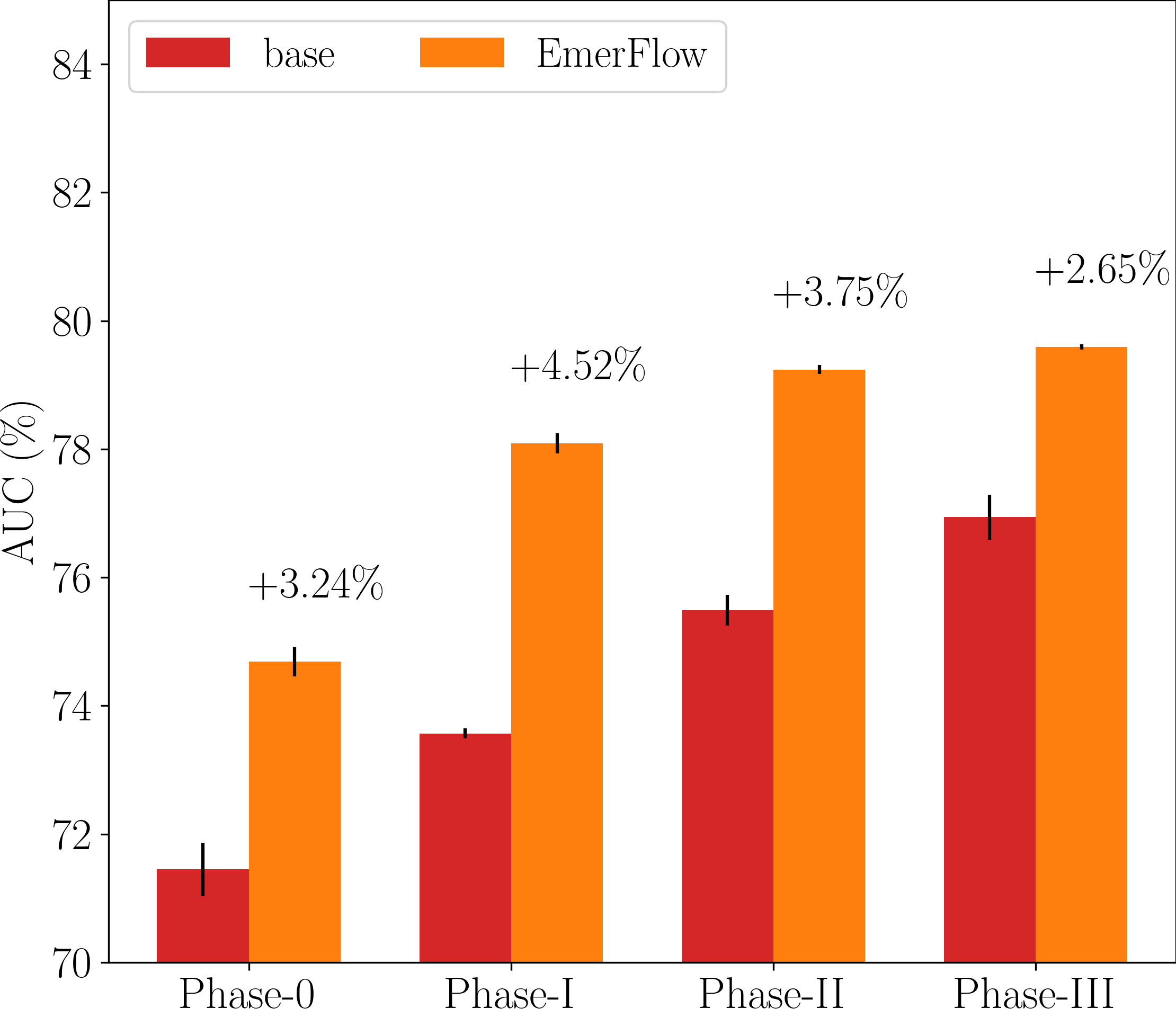}
	}
	\subfigure[AutoInt.]{
		\includegraphics[width=0.23\textwidth]{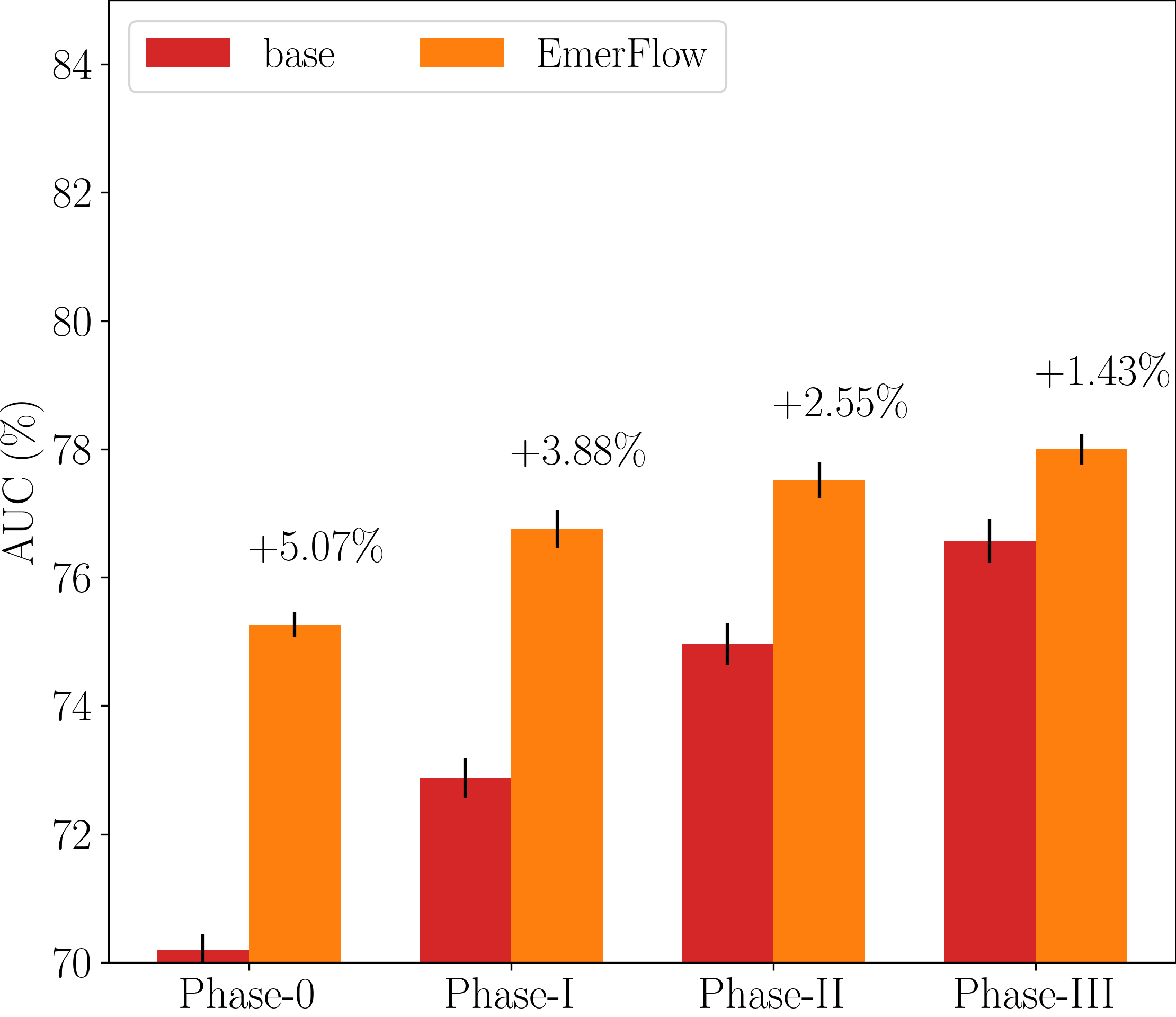}
	}
	\caption{The AUC score of \TheName{} and the base backbone models. The percentage on orange bar means the how much \TheName{} improves AUC on the base of backbone.}
	\label{fig:different-backbone}
\end{figure*}

\textbf{Influence of Different Backbones.}
We further implement \TheName{} on multiple backbone models and compare its performance with their standalone versions.  
As illustrated in Figure~\ref{fig:different-backbone}, \TheName{} consistently improves AUC over the base backbones, demonstrating its plug-and-play nature and effectiveness across diverse architectures.

\begin{figure*}[htbp]
	\centering
	\subfigure[\emph{Case1}.]{
		\includegraphics[width=0.31\textwidth]{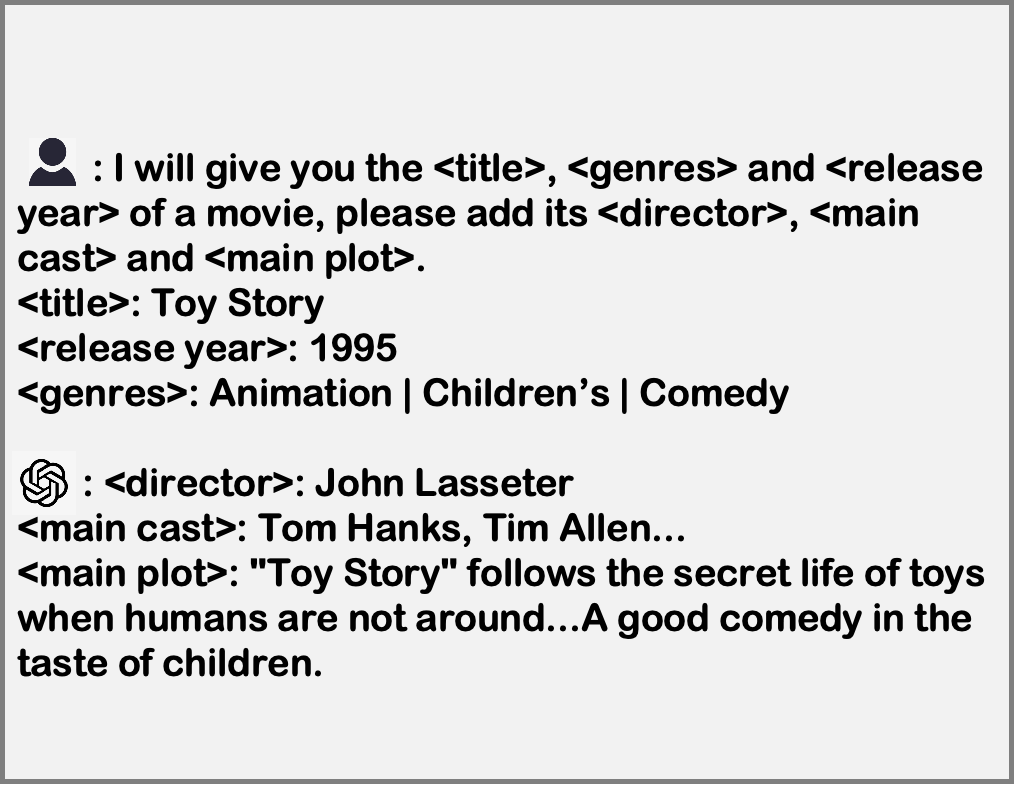}
	}
	\subfigure[\emph{Case2}.]{
		\includegraphics[width=0.31\textwidth]{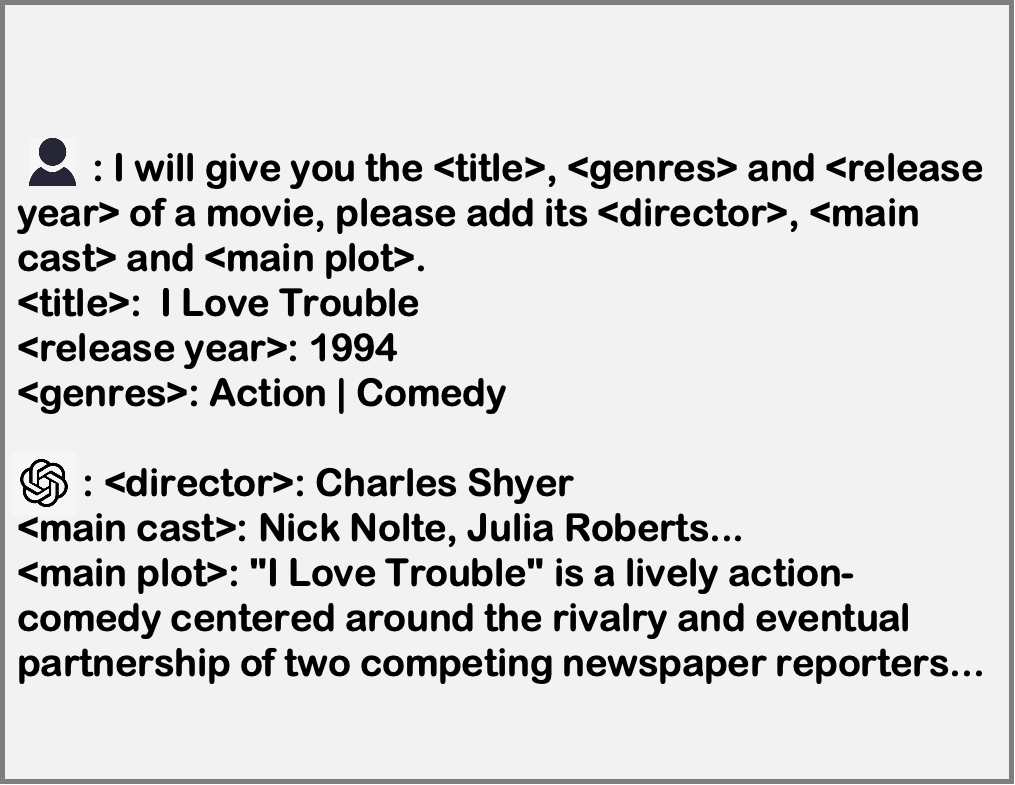}
	}
	\subfigure[\emph{Case3}.]{
		\includegraphics[width=0.31\textwidth]{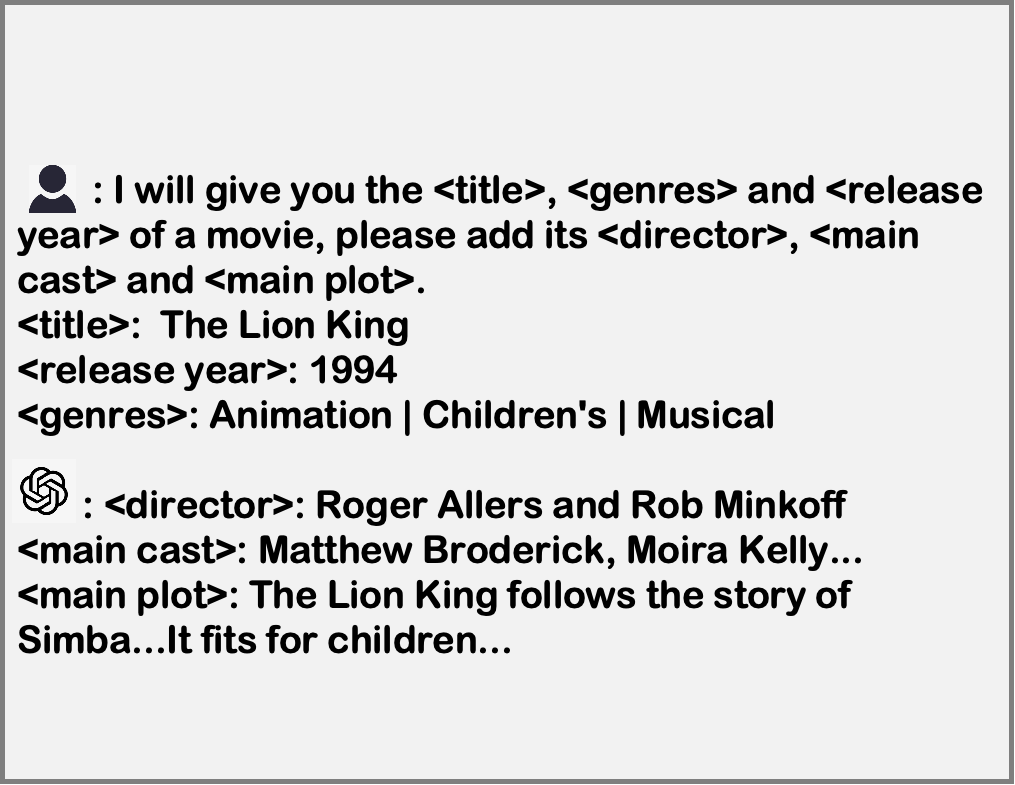}
	}
	\caption{When the LLM supplement the movie's main plot, the title and genres are usually in it.}
	\label{fig:case}
	\vspace{-10pt}
\end{figure*}

\subsection{Case Study and Qualitative Analysis}

\begin{figure}[htbp]
	\vspace{-10pt}
	\centering
	\includegraphics[width=0.4\textwidth]{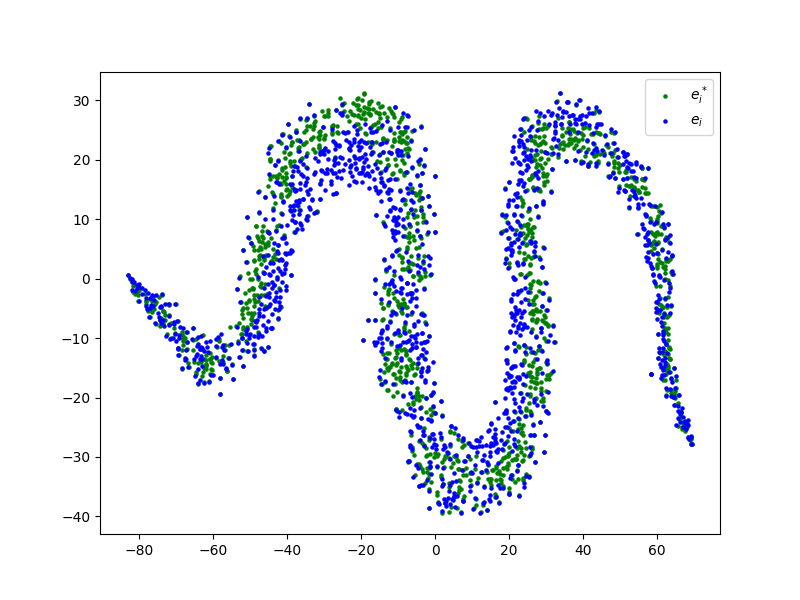}
	\caption{t-SNE of item representations learned by backbones and \TheName{} for old items.}
	\label{fig:tsne-movielens}
	\vspace{-10pt}
\end{figure}

\textbf{Visualizing Learned Representations.}
We conduct a case study on \emph{MovieLens} to illustrate that \TheName{} can generate item embeddings closely matching those of fully trained old items.  
For each old item $v_i$, let $\be_i$ denote the embedding generated by \TheName{} and $\be_i^*$ denote the embedding obtained by fully training the backbone with abundant interactions.  
We visualize both embeddings in Figure~\ref{fig:tsne-movielens}.  
The results show that the embeddings produced by \TheName{} effectively capture inter-item relationships and align well with those learned from sufficient supervision.

\textbf{Feature Overlap between LLM and Dataset.}
To further explain the comparable performance of \textbf{w/o $\cF_i$} and \TheName{} observed in the ablation study, we analyze the semantic overlap between LLM-provided and dataset-provided features.  
Figure~\ref{fig:case} provides illustrative examples.  
When the LLM describes a movie’s main plot, it often includes the movie title and genres—attributes already contained in the dataset.  
Therefore, even when \textbf{w/o $\cF_i$} encodes only LLM-generated features, it implicitly captures most of the original item information.  
This qualitative observation supports our hypothesis that the LLM augmentation inherently subsumes the core dataset features, explaining the similar quantitative results.

\section{Conclusion}

This paper introduced EmerFlow, a new framework that leverages Large Language Models (LLMs) to enhance the recommendation of newly emerging items with limited interactions. By enriching raw item features and aligning them with established models through a meta-learning approach, EmerFlow addresses the dynamic nature of data accrual that traditional methods often overlook. 

Our experiments across various domains, including movies and pharmaceuticals, have convincingly demonstrated that EmerFlow outperforms conventional recommendation methods. These results underscore its effectiveness in environments where data is not only sparse but also accumulates dynamically, making it highly relevant for industries facing rapid changes in item popularity and consumer interests.

However, despite its strengths, EmerFlow's reliance on LLMs for feature enrichment can pose challenges. The complexity introduced by these models may lead to difficulties in interpretation and troubleshooting, particularly when unexpected data patterns emerge or when anomalies need to be addressed. Additionally, the computational demands associated with training and deploying LLMs could limit scalability in resource-constrained scenarios.

Looking forward, the potential to refine EmerFlow is vast. Future work could focus on increasing the framework's efficiency and adaptability, particularly during later phases of interaction accrual. Investigating more advanced meta-learning algorithms and exploring more dynamic data integration strategies could further enhance its performance and applicability.

\section*{Acknowledgements}
Y. Wang is sponsored by Beijing Nova Program.

%\clearpage
\bibliographystyle{unsrt} 
\bibliography{coldllm}

\end{document}